\documentclass[12pt]{article}
\RequirePackage[authoryear]{natbib}
\usepackage[symbol]{footmisc}

\RequirePackage{amsthm,amsmath,amsfonts,amssymb}
\RequirePackage[colorlinks,citecolor=blue,urlcolor=blue]{hyperref}
\RequirePackage{graphicx}

\usepackage{float}
\usepackage{enumitem}
\usepackage{url}
\usepackage{bbm}
\usepackage[nottoc,numbib]{tocbibind}
\usepackage{booktabs,caption,subcaption}
\usepackage[flushleft]{threeparttable}
\usepackage{multirow}

\usepackage[pdftex,dvipsnames]{xcolor}
\usepackage{comment}

\usepackage[margin=1in]{geometry}

\usepackage{macros}

\theoremstyle{definition}

\begin{document}

%\jmlrheading{23}{2022}{1-\pageref{LastPage}}{1/21; Revised 5/22}{9/22}{21-0000}{}

\title{\bf Improving Event Time Prediction by Learning to Partition the Event Time Space}
 \author{Jimmy Hickey$^{1,}$\footnote{Corresponding author: \href{jhickey@ncsu.edu}{jhickey@ncsu.edu}} , Ricardo Henao$^{2}$, Daniel Wojdyla$^{3}$,\\\vspace{.2cm} Michael Pencina$^{3,4,5}$, Matthew M. Engelhard$^{4,5}$\hspace{.2cm}\\ 
  $^{1}$\textit{\small Department of Statistics, North Carolina State University}\\
  $^{2}$\textit{\small Biological, Environmental Sciences and Engineering}, \\ \textit{\small King Abdullah University of Science and Technology}\\
  $^{3}$\textit{\small Duke Clinical Research Institute}\\
  $^{4}$\textit{\small Department of Biostatistics and Bioinformatics, Duke University School of Medicine}\\
  $^{5}$\textit{\small Duke AI Health}\\
  }
  \date{}
 \maketitle
 
 \bigskip
\begin{abstract}
   Recently developed survival analysis methods improve upon existing approaches by predicting the probability of event occurrence in each of a number pre-specified (discrete) time intervals. By avoiding placing strong parametric assumptions on the event density, this approach tends to improve prediction performance, particularly when data are plentiful. However, in clinical settings with limited available data, it is often preferable to judiciously partition the event time space into a limited number of intervals well suited to the prediction task at hand. In this work, we develop a method to learn from data a set of cut points defining such a partition. We show that in two simulated datasets, we are able to recover intervals that match the underlying generative model. We then demonstrate improved prediction performance on three real-world observational datasets, including a large, newly harmonized stroke risk prediction dataset. Finally, we argue that our approach facilitates clinical decision-making by suggesting time intervals that are most appropriate for each task, in the sense that they facilitate more accurate risk prediction.\end{abstract}

\noindent%
{\it Keywords:} survival analysis, time-to-event, interval prediction, limited data

\section{Introduction}
Time to event modeling, also called survival analysis, is ubiquitous throughout clinical medicine as well as in many other fields concerned with predicting risk of events of interest (\textit{e.g.}, clinical outcomes) based on available features (\textit{e.g.}, patient characteristics). Traditional approaches include the well-known Cox proportional hazards (Cox-PH) model \citep{cox1972regression}, in which features modulate a baseline hazard rate; and the accelerated failure time (AFT) model \citep{wei1992accelerated} model, in which features accelerate or decelerate a learned, parametric event time density.

Recently developed methods have focused on $a$) allowing effects of features on the hazard rate or event time density to be non-linear and flexible \citep{katzman2018deepsurv, ranganath2016deep, kvamme2019time, miscouridou2018deep}; and $b$) also allowing greater flexibility in the form of the event time density itself via approaches that \textit{discretize} time, then predict the probability of event occurrence in each resulting \textit{time interval} \citep{yu2011learning, lee2018deephit, ren_deep_2018, tjandra2021hierarchical, engelhard2022disentangling}.

The prognostic information provided by these models often has direct and significant impact on stakeholder decision-making. In a clinical setting, for example, information about risk within a particular time interval might influence providers' or patients' decisions about whether to pursue treatment, or which specific treatment to pursue. It is therefore critical not only that predictions are accurate, but also that they are easily interpretable by stakeholders who wish to integrate them in decision-making. The predictions of a Cox-PH model might be presented to stakeholders as relative hazards, for instance, whereas it is natural to present the predictions of more recent models as the probability of event occurrence in a specific time interval of interest.

Importantly, however, decisions about these intervals made during model development -- in other words, choices about the number and placement of \textit{cut points} used to discretize the event time space -- can have substantial impact on interpretability as well as performance. Equipped with unlimited data, we might use a large number of cut points to divide the timeline into tiny intervals; this would then allow us to summarize risk over an arbitrary time period of interest by combining predictions across all the intervals that comprise that period. However, the amount of data required to accurately estimate risk in each interval increases as the number of intervals increases, making this approach impractical even for large observational datasets. Equipped with unlimited time, on the other hand, we might present risk in a format most relevant to a particular patient, or to the decision at hand. Again, however, practical considerations typically require us to instead summarize risk over a consistent, limited number of time-frames (\textit{e.g.}, 10-year risk, 5-year risk). In some cases a particular discretization is most actionable given the clinical context, but in others the choice is arbitrary, and it would be preferable to identify a discretization that facilitates more accurate prediction.

To illustrate the problem more concretely, consider the following example from the maternal health setting, which partly motivated this work. Patients with preeclampsia and gestational hypertension have substantially increased risk of postpartum cardiovascular events \citep{meng2022maternal}, but this risk could be mitigated by regular monitoring (\textit{e.g.,} increased visits) of high-risk patients in the months after delivery. When developing a monitoring strategy, it is important to determine not only (a) which patients are at highest risk, but also (b) how long monitoring should take place; yet we have limited data available for learning because the outcome rates are low.

Our goal, therefore, is to develop a principled, data-driven approach to answer both of these questions. Specifically, we wish to develop a method that providers can use to identify time intervals that are optimal when \textit{understanding} risk, for example to design an intervention or monitoring strategy, as well as when \textit{reporting} risk to patients. At the same time, we wish to retain the substantial advantages and flexibility of other recently developed approaches, including their lack of strong parametric assumptions about the form of the event density.

We begin by recasting learning from \textit{discrete} survival times as learning from \textit{continuous} survival times under the assumption that the density is piecewise constant; and then formulate a smooth relaxation of this piecewise constant density that allows cut points (\textit{i.e.}, interval boundaries) to be learned by gradient-based optimization methods. We then present our learning procedure and results of experiments with two simulated and three real datasets -- including a newly harmonized stroke risk prediction dataset that pools data across three large cohorts -- that illustrates the effectiveness and potential clinical relevance of our approach.

Our performance evaluation focuses on comparing our method to its state of the art alternative, namely, discrete-time, neural network-based risk prediction over fixed-length intervals.

In summary, our contributions are as follows:
\begin{itemize}
    \item Present a novel model and associated learning procedure to learn an optimal event time partition from data rather than fixing it \textit{a priori}.
    \item Present simulation results illustrating effective learning of cut points that are consistent with the true, underlying generative model.
    \item Demonstrate improved prediction performance across three real datasets, including two clinical datasets.
    \item Identify clinically meaningful risk cut points illustrating the potential of the approach to provide improved prognostic information.
\end{itemize}

\section{Methods}
\label{sec: Methods}

\subsection{Setup and Notation}
%
% exposition about what are cut points and why do we need them
Consider a time-to-event outcome where each observation is represented by the triplet $\{\bX, Y, S\}$, where $\bX \in \mathcal{X} \in \mathbb{R}^{p}$ is a $p$-dimensional feature vector, $Y \in (0, T_{\text{max}}]$ is an observed event time over a finite time horizon, and $S\in \{0,1\}$ indicates whether $Y$ is a right-censoring time ($S=0$) or an event time ($S=1$).
The observed time $Y$ is the minimum of the event time $T$ and the right-censoring time $U$, {\em i.e.}, $Y=\min(T, U)$, and $S=\mathbbm{1}(T<U)$, where the indicator function $\mathbbm{1}(\cdot)$ is 1 when the argument is true and 0 otherwise.

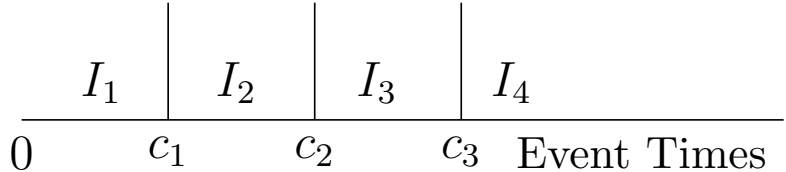
\begin{figure}[t]
\centering
% scale tikz picture
\resizebox{0.65\linewidth}{!}{%
\begin{tikzpicture}
\draw[black] (0,0) node[below]{0}--
    (1,0) node[above left]{$I_{1}$} --
    (2.15,0) node[above left]{$I_{2}$} --
    (3.35,0) node[above left]{$I_{3}$} --
    (4.5,0) node[above left]{$I_{4}$} --
    (6.5,0) node[below left]{\small Event Times} ;
\draw[black](1.25,0) node[below]{$c_{1}$} -- (1.25,1)  ;
\draw[black](2.5,0) node[below]{$c_{2}$} -- (2.5,1) ;
\draw[black](3.75,0) node[below]{$c_{3}$} -- (3.75,1) ;
\end{tikzpicture}
} % end scale
% \captionsetup{width=.85\linewidth}
\caption{
The event time space partitioned by three cut points into four intervals.
}
\label{figure: intervals}
\end{figure}

We consider possible sequences of $M$ \textit{cut points} $C=\{c_j\}_{j=1}^M$, where $0 = c_0 < c_1 < \cdots < c_M < c_{M+1} = T_{\text{max}}$, that partition the event time space, $(0, T_{\text{max}}]$, into the intervals $I_1, \ldots, I_{M+1}$, where $I_j=(c_{j-1},c_j]$.
Figure~\ref{figure: intervals} provides an example of the event time space partitioned into four intervals: $I_{1} = (0, c_{1}]$, $I_{2}=(c_{1}, c_{2}]$, $I_{3} = (c_{2}, c_{3}]$, and $I_{4} = (c_{3}, T_{\text{max}}]$.
Given such a partition, we introduce an auxiliary random variable $Z \in \{1, \ldots, M + 1\}$ that indicates which interval contains $T$, {\em i.e.}, $Z=j \iff t \in I_j$.

\subsection{Piecewise Constant Density}
\label{sec: piecewise}
We begin by considering learning with fixed cut points, which is currently the predominant approach.
For example, \citet{lee2018deephit} and other recently-developed methods \citep{ren_deep_2018, tjandra2021hierarchical, engelhard2022disentangling} use fixed cut points to discretize time in order to avoid placing restrictive, parametric assumptions on the form of the event time density. Instead, the density is restricted to be piecewise constant according to the intervals defined by the cut points.
The cut points themselves might be evenly spaced in time, or alternatively they might be evenly spaced across the observed or estimated event time distribution, {\em e.g.}, via empirical quantiles.
The goal of learning is then to estimate $P(Z|\bX)$, the conditional probability that $T$ will fall in each of the pre-defined intervals, rather than $p(T|\bX)$, the conditional density of $T$. Typically $T$ is discretized to $Z$ {\em a priori}.

However, it is not possible to \textit{learn} the cut points $C$ with this approach, because $Z$ depends on $C$ in addition to $T$.
To see this, consider the value of $Z$ associated with an observed time $t \in (0, T_{\text{max}}]$ under the binary partition defined by the single cut point $c_1$.
If we choose $c_1 \ge t$, then $t \in (0, c_1]$, therefore $Z=1$; but for $c_1 < t$, we have $t \in (c_1, T_{\text{max}}]$, therefore $Z=2$.

To circumvent this limitation, we note that estimating $P(Z|\bX)$ is equivalent to estimating $p(T|\bX)$ with the following piecewise constant model, which supposes $p(T|\bX)$ has uniform density over each interval $I_j$:
\begin{equation}
\label{eq:piecewise}
    \hat{p}(t|\bx)=\sum_{j=1}^{M+1} p_\phi(z_j|\bx)\frac{\mathbbm{1}_{I_j}(t)}{|I_j|},
\end{equation}
where $\mathbbm{1}_{I_j}(\cdot)$ is the indicator function associated with the interval $I_j$, and $\phi$ parameterizes our model of $P(Z|\bX)$.
Importantly, we must normalize by $|I_j|$, the length of $I_j$, to ensure $\int_{(0, T_\text{max}]}\hat{p}(t|\bx)=1$ and $\int_{I_j}\hat{p}(t|\bx)=p_\phi(z_j|\bx)$. Subsequently we will see how this view facilitates learning of the cut points $C$.
%
% Our goal is to find a partition $C$ that maximizes the expected information gain associated associated with observing $\bX$; in other words, we wish to maximize the mutual information $H(\mathcal{I}_C) - H(\mathcal{I}_C | \bX)$.
%\rh{I changed the interval covering 0 as open. By definition, $Y$ and $T$ cannot be 0.}

\subsection{Smooth Relaxation of Piecewise Density}
The parameters $\phi$ of our model for $Z$ can be learned directly from equation~\eqref{eq:piecewise}. However, our goal is to learn not only $\phi$ but also $C$, the specific partition that allows our model to best approximate $p(T|\bX)$ across a given dataset. Unfortunately, \eqref{eq:piecewise} cannot be optimized with respect to $C$ via gradient-based methods.
This is because the indicator function $\mathbbm{1}_{I_j}(\cdot)$ implicitly depends on $C$, and is discontinuous whenever a cut point is equal to an observed event time.

To illustrate, consider learning a single cut point $c_1$ while holding the parameters $\phi$ fixed.
%\rh{what is the $*$ in $\phi_*$? same for $t_*$ below}
For small $\varepsilon$ such that $0 < \varepsilon < t$, where $t$ is an observed event time associated with covariates $\bx$, suppose the cut point $c_1=t + \varepsilon$ is just after the observed event time.
In this case, we have $t \in I_1$, therefore $\mathbbm{1}_{I_1}(t)=1$ and $\mathbbm{1}_{I_2}(t)=0$, and consequently $\hat{p}(t | \bx) = p_\phi(z_1 | \bx)/|I_1|$. 
%\rh{I still do not get your notation, if $t_* \in I_2$ shouldn't $p(t_* \in I_2; \bX, c_1, \phi_*)$ be 1 by definition? I understand that this is your likelihood, but there is some circularity to the way it is defined.}
On the other hand, suppose the cut point $c_1=t - \varepsilon$ is just before the observed event time.
In this case, we have $t \in I_2$, therefore $\mathbbm{1}_{I_1}(t)=0$ and $\mathbbm{1}_{I_2}(t)=1$, and consequently, $\hat{p}(t | \bx) = p_\phi(z_2 | \bx)/|I_2|$.
Thus, for any non-trivial model $p_\phi$ for which $p_\phi(z_1 | \bx) \ne p_\phi(z_2 | \bx)$, equation \eqref{eq:piecewise} is discontinuous at $c_1=t$.
This argument readily generalizes to all cut points.

To smooth this discontinuity and allow gradient-based optimization, we replace the indicator function $\mathbbm{1}_{I_j}(t)$ in \eqref{eq:piecewise} with the smooth approximation $\sigma((t - c_{j-1})/\tau)*\sigma((c_j - t)/\tau)$, where $\sigma(z) = (1 + e^{-z})^{-1}$ is the sigmoid function.
The temperature $\tau$ is a hyperparameter of the model that should be tuned based on the scale of the observed event times. 
%\rh{are you sure the formulation is correct, shouldn't it be something like $\sigma(\cdot)(1- \sigma(\cdot))$?}

This results in the following relaxed model:

\begin{equation}
\label{eq:smoothed}
    \hat{p}(t|\bx)=\sum_{j=1}^{M+1} p_\phi(z_j|\bx)\frac{\sigma(\frac{t - c_{j-1}}{\tau})\sigma(\frac{c_j - t}{\tau})}{|I_j|},
\end{equation}
which is approximately piecewise constant for $\tau \ll T_\text{max}$, yet differentiable everywhere with respect to $C$ and thus suitable for gradient-based optimization.
%
%\rh{write the RHS.}
%

% We model this timeline partition as a mixture distribution with $M + 1$ components corresponding to the $M + 1$ intervals $I_j$. We assume that given $T \in I_j$, there is no further dependence of $T$ on $X$, {\em i.e.}, $T \ind \bX \mid T \in I_j$; and further that all $t \in I_j$ are equally likely, {\em i.e.}, $T \sim \text{Unif}(c_{j-1}, c_j) \mid T \in I_j$.
% A distinguishing feature of this model is that individual components are non-overlapping, so that $\forall j,k$ where $k \ne j$, $P(t \in I_j) > 0$ implies $P(t \in I_k) = 0$.  We will later relax this assumption to allow gradient-based optimization.
% We may therefore write $p(T | \bX; \theta)$ as follows, where $\theta = \{C, \phi\}$ comprises the learnable cut points $C$ along with the parameters $\phi$ of the model predicting the probability of membership in each of the intervals $I_j$.
% Specifically,
% %
% \begin{equation}
% \label{eq:lik}
%     p(T | \bX; \theta) = \sum_{j=1}^{M+1} \frac{\mathbbm{1}(T \in I_j)}{|I_j|} p(T \in I_j | \bX ; \theta),
% \end{equation}
% %
% where \rh{... Note that this notation is somewhat weird because the $\mathbbm{1}(T \in I_j)$ is only 1 if $T$ is in the interval, which will render $p(T \in I_j | \bX ; \theta)=1$ by definition, and $|I_j|$ is either the cardinality or the size of the interval. The former does not make a lot of sense and the latter normalizes what exactly (relative to the numerator)?}
%
%$\mathbbm{1}(\cdot)$ is 1 when $(\cdot)$ is true and 0 otherwise.

\subsection{Learning Procedure}
%
% \begin{equation}
%     \sum_{j=1}^{M+1} p(T | T \in I_j; \theta)p(T \in I_j | \bX ; \theta)
% \end{equation}
%
%We begin with a single level before finally extending to multiple levels.
Under the common assumption of non-informative right-censoring, we may ignore the censoring density and optimize $\hat{p}(y,s | \bx; \theta)$, where $\theta = \{\phi, C\}$, over the observed data $\mathcal{D}=\{\bx_i, y_i, s_i\}_{i=1}^N$
as follows:
\begin{align}
\label{eq:optimize}
    \theta & = \argmax_{\theta}  \sum_{i}^{N} \big[s_{i} \log \hat{p}(t_{i}|\bx; \theta) \\
    & \hspace{2cm} + (1 - s_{i})\log \hat{P}(t_{i}>y_{i}|\bx_{i}; \theta)\big], \notag
\end{align}
where $\hat{P}(t_{i}>y_{i}|\bx_{i}; \theta) = 1 - \int_0^T \hat{p}(\tau| \bx_{i}; \theta)$ is the survival function associated with $\hat{p}(t_{i}|\bx_{i}; \theta)$ for observation $i$.

However, optimizing equation~\eqref{eq:smoothed} alone can result in degenerate solutions in which cut points become arbitrarily close together or even coincide. In the extreme case, it is possible to have $I_j=(0, T_\text{max}]$ for a particular $j \in \{1, M+1\}$, resulting in the trivial model in which $p_\phi(z|\bx)$ places all mass on $z_j$.

% It is therefore critical to balance optimizing equation~\eqref{eq:smoothed} versus ensuring that $p_\phi(z|\bx)$ is non-trivial.
% We accomplish this by incorporating the entropy $H(p_\phi(z|\bx))$ of $p_\phi(z|\bx)$ as a regularization term with associated hyperparameter $\lambda_1$ in our optimization procedure.
% As described in subsequent sections, $H(p_\phi(z|\bx))$ is also emphasized in our evaluation, which focuses on the trade-off between prediction performance and the average entropy of our predictions.

It is therefore critical to balance optimizing equation~\eqref{eq:smoothed} versus ensuring that $p_\phi(z|\bx)$ is non-trivial.
We accomplish this by incorporating a regularization term, $H(p_\phi(z|\bx))$, with associated hyperparameter $\lambda_1$ in our optimization procedure.
We use a scaled $\text{Beta}(1.5, 1.5)$ distribution on each cut point.
For example, suppose there are three cut points $c_{1} < c_{2}^{\star} < c_{3}$ where $c_{2}^{\star}$ is the newly proposed value for the middle cut point $c_{2}$.
We scale the value of the cut point to find its location relative to the cut points near it: $c_{2, \text{scaled}}^{\star} = ( c_{2}^{\star} - c_{1} ) / (c_{3} - c_{1})$.
The final regularization value is the PDF value of $c_{2, \text{scaled}}^{\star}$ evaluated over a $\text{Beta}(1.5, 1.5)$ distribution.
This regularization term encourages cut points to be near the center of their two surrounding cut points.

We may then optimize $\theta$ over $\mathcal{D}$ by choosing $\theta=\argmin_\theta \sum_\mathcal{D} \mathcal{L(\theta)}$, where $\mathcal{L(\theta)}$ is defined as follows:
%
%  + \lambda_2 R(\theta)
%
\begin{equation}
\label{eq:loss}
    \mathcal{L}(\theta) = -\log \hat{p}(y, s \mid \bx; \theta) - \lambda_1 H(p_\phi) .
\end{equation}
Here the first term is the negative log likelihood in \eqref{eq:optimize} and the second is our beta based regularizer.
% Finally, we reinforce learning at all levels of the hierarchy, jointly, by defining $\mathcal{L}^{\ell_1}, \mathcal{L}^{\ell_2}, \ldots$ corresponding to each level $\ell_k$, $k=1,\ldots,K$, of the cut points.
Our learning procedure then becomes:
\begin{equation}
\label{eq:hierloss}
    \theta=\argmin_\theta \sum_\mathcal{D} \mathcal{L}(\theta) + \lambda_2 R(\theta) ,
\end{equation}
where we have included an additional regularization term $R(\cdot)$ (\textit{e.g.}, $L_2$-regularization) along with an associated hyperparameter $\lambda_2$ to control for overfitting.

\section{Implementation Details}
\label{sec: implementation details}

\subsection{Baseline Model: Discrete-Time Neural Network}
\label{sec: baseline model}
We compare our model to a discrete-time neural network baseline that is identical to the proposed model, except the cut points (and corresponding intervals) are initialized based on the observed outcomes and remain fixed when learning the classifier. This approach, hereafter called the \textit{DTNN Baseline}, was popularized by DeepHit \citep{lee2018deephit} and is currently the predominant approach.

We initialize the DTNN Baseline model's cut points to be evenly spaced on the percentiles of the empirical Kaplan-Meier curve of the observed outcomes; for example, if there are three cut points then they would be placed at the time points associated with the $25^{\text{th}}$, $50^{\text{th}}$, and $75^{\text{th}}$ percentiles on the estimated Kaplan-Meier curve. 
% Another approach is to place them evenly across the event times; for example, if the maximum event time is 200, then the would be placed at $t = 50, 100, 150$.
With these cut points fixed, we then build a model predicting the probability that the patient will experience the outcome in each interval.
Note that this differs from our method in which we also consider the cut points themselves as parameters.
The DTNN Baseline classification model learns the probability of each observation being in each of the pre-defined intervals.
In the notation of Section \ref{sec: piecewise}, the DTNN Baseline approach learns only the model parameters, $\phi_{\text{baseline}}$, whereas our hierarchical approach learns both model parameters $\phi$ and the cut points $C$.
Importantly, we search the same grid of hyperparameters for the DTNN Baseline model as for our proposed method.

\subsection{Performance Quantification}
\label{sec: perfomance measurement}

With simulated data we were able to judge the correctness of the estimated cut points by their proximity to the true cut points used in the data generation process. 
For the real data we do not know the true cut point values and thus need other metrics to judge our model's performance.

\noindent
\textbf{Time-Dependent Concordance Index (CI)}
Since we consider a time-to-event outcome with censored observations rather than a regression or classification outcome, standard metrics such as root mean square error and area under the receiver operating characteristic are insufficient to capture the prediction performance of our method.
Initially developed by \cite{harrell1984regression}, the concordance index (CI) measures how well predicted event times match the order of the true event times.
However, both our proposed approach and the DTNN Baseline predict discrete interval membership instead of continuous event times, and the ordering of predicted risk can change over time. To properly account for these characteristics, we use a discrete-time implementation of the time-dependent concordance index developed by \cite{antolini2005time}.
This metric compares model-predicted risk at observed failure times to the model-predicted risks \textit{at that time} for other individuals known to have later failure times.
Pairs of individuals are only considered if ($a$) both failure times are known (neither are censored), or ($b$) one failure time is known to have occurred before the censoring time of the other. 

% \noindent
% \textbf{Entropy}
% We also consider the entropy defined in \cite{shannon1948mathematical} which measures relative size of each interval.
% A high entropy value indicates that each interval has a similar number of observations.
% An extreme example of low entropy is if all of the cut points move to the maximum time point.
% This would result in all of the observations being in one interval while the other intervals would be empty; this would make the prediction task trivial.

\noindent
\textbf{AUC at last cut point}
The Area Under the Receiver Operating Characteristic Curve (AUC) is a common metric to evaluate predictive performance for a binary outcome.
To adapt this to our method, we focus on the AUC at the last cut point.
That is, we are interested in determining if our method is able to predict whether an event happens before or after the final cut point.
This is especially relevant for data sets with high amounts of censoring at the end of the study.
The cases are all observations that experienced an event prior to the final cut point and the controls are all observations with an observed time (either an event or censored) after the final cut point.
Notice that observations that are censored prior to the last cut point are omitted from this metric.

\noindent
\textbf{Calibration slope and intercept}
We also consider the calibration slope and intercept as described by \citet{crowson2016assessing}, which quantify the degree to which model-predicted probabilities accurately estimate true event probabilities, as determined based on observed event rates.
A well calibrated model will have a calibration slope near 1 and a calibration intercept near 0.

\subsection{Hyperparameter Tuning}
\label{sec: hyperparameter tuning}
Our method is flexible, allowing for any number of cut points.
In our simulation examples we will know exactly how many cut points were used to generate the data; however, this is not the case for the real data experiments.
So we use 3, 5, and 10 cut points.
We use a two layer neural network as our predictive model.
The first layer has input dimension $p$ based on the feature dimension of the data and output dimension $h$, for which we explore values of 32, 128, and 512.
This is then connected by a Rectified Linear Unit activation function to another layer with input dimension $h$ and output dimension.
These networks are optimized using Adam \citep{kingma2014adam} with a learning rate of 0.01 and weight decay values between 0.0001 and 0.1.
We vary the strength of the regularization on the cut points $\lambda_{1}$ from values in the range of 0.1 to 20 and use a mini-batch size of 64 for the training data.

\begin{figure}[t]
     \centering
     \includegraphics[width= 0.8\textwidth]{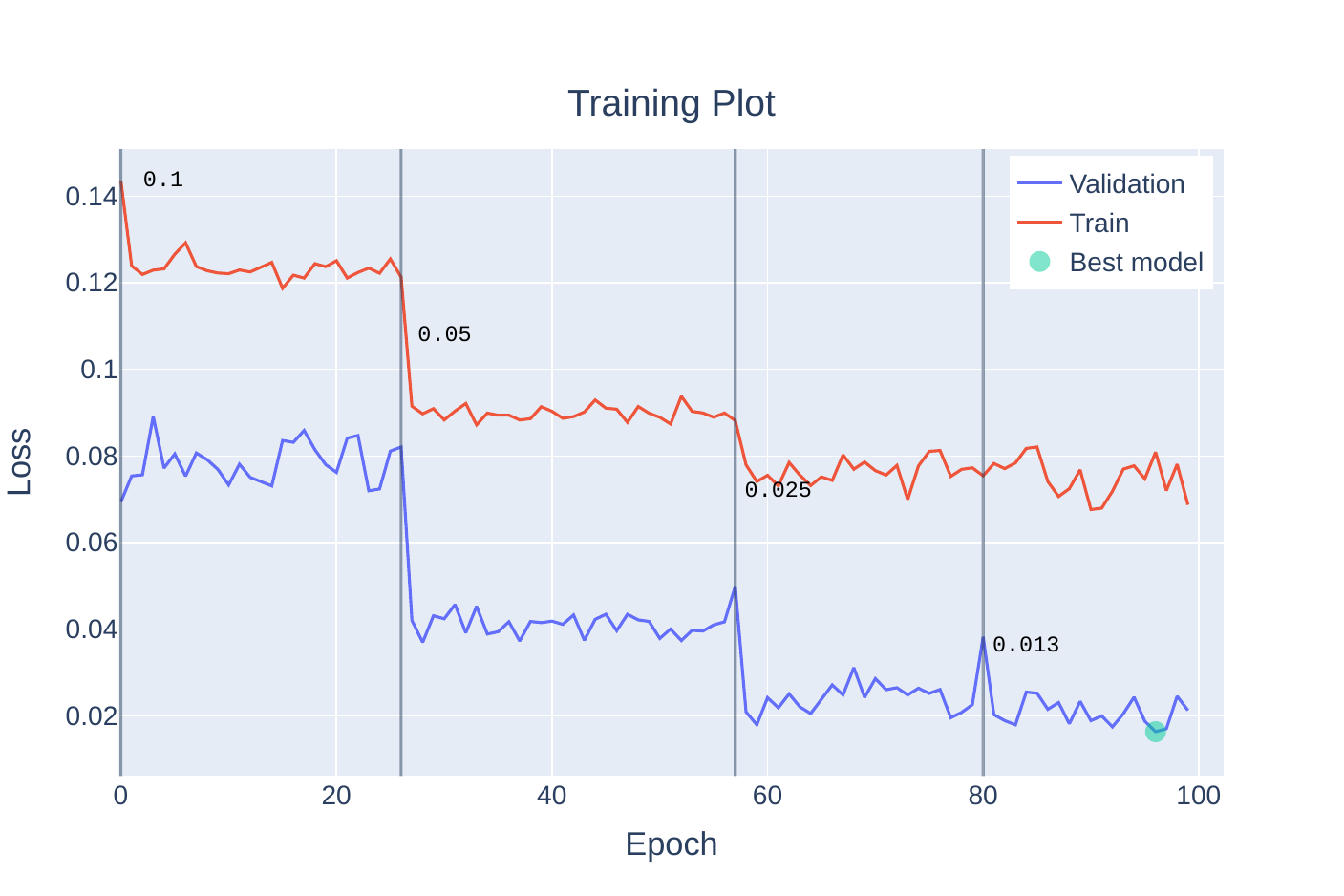}
     \caption{A training plot of the training and validation loss at each epoch for the two interval simulation example. The vertical lines represent drops in sigmoid temperature $\tau$ and the accompanying new value of $\tau$.}
     \label{fig: training plot}
\end{figure}

During training, we initially set the sigmoid temperature used in our smooth approximation (see equation~\eqref{eq:smoothed}) to a value $\tau = 0.1$, then lower it when the loss stops changing significantly between epochs.
Lowering the temperature reduces the degree of smoothing and sharpens the boundaries between intervals defined by each cut point.
Figure \ref{fig: training plot} shows an example training plot where the temperature drops after multiple epochs with no improvement in the validation loss.
It is clear that this drop then leads to an improvement in both training and validation loss.

We perform a grid search over the hyper parameters, testing every combination of output dimension, weight decay, and regularization strength.
The evaluation process to compare hyperparameters is described in Section \ref{sec: perfomance measurement}.
We train each network for 250 epochs.

To evaluate performance we perform five-fold cross validation.
We randomly partition the data into training (75\%), validation (15\%), and test (10\%) sets.
For each set of hyperparameters we perform this partition five times, using the training sets for learning the model parameters.
We then calculate average performance metrics metrics on the out of sample validation sets.
Only the model with the best average validation set performance is then applied to the corresponding, yet unseen, test sets.
We report the average and standard deviations of the performance metrics calculated across the folds on the test sets.
Through this general cross-validation strategy, we are able to find the hyperparameter setting that performs the best on out of sample data from the hyperparameters tested.
We report the mean and standard deviation of each metric across the folds.

We perform the same parameter search and evaluation to find the best DTNN Baseline model as described in Section \ref{sec: baseline model}.
We compare the metrics of our best model to that of the best baseline model.
We report the CI and AUC for both methods calculated on the unseen test set.

\section{Simulation Examples}
\label{sec: simulation study}

\subsection{Learning Two Intervals}
\label{sec: 1d}
\begin{figure*}[ht]
     \centering
     \begin{subfigure}[b]{0.27\textwidth}
         \centering
         \includegraphics[width=\textwidth]{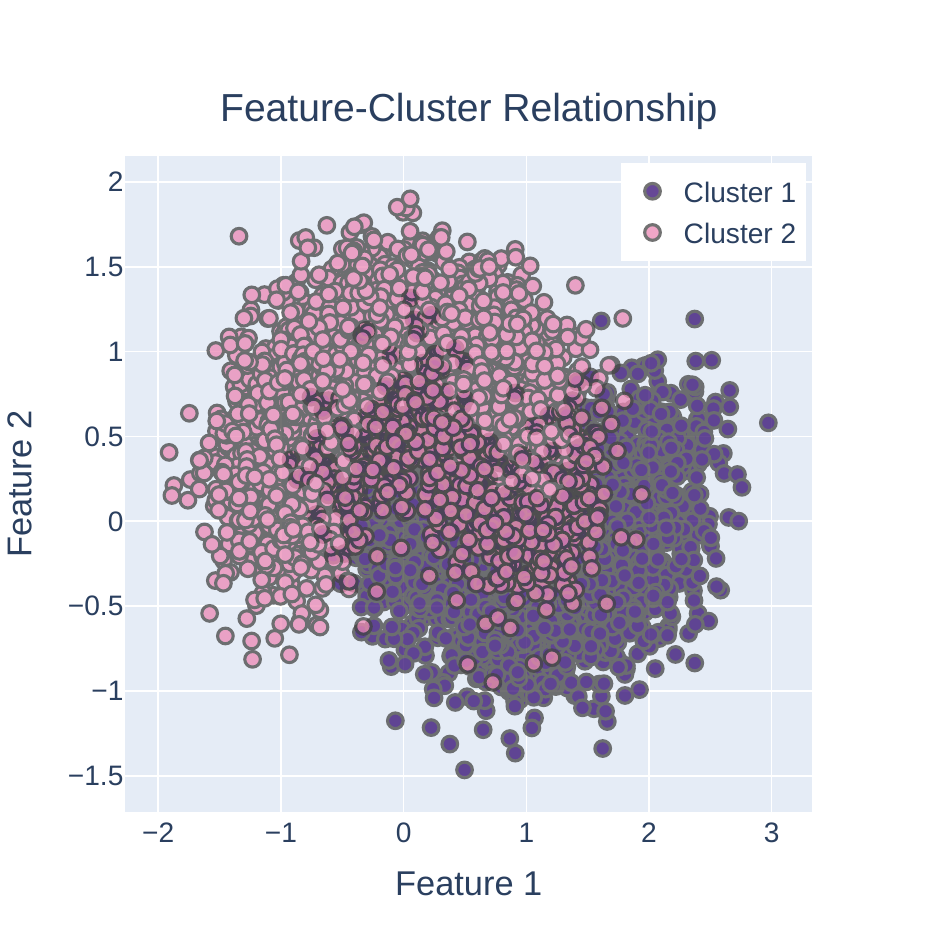}
         \caption{The noisy, nonlinear feature-cluster relationship.}
         \label{fig: 1d feature clusters}
     \end{subfigure}
     \hfill
     \begin{subfigure}[b]{0.27\textwidth}
         \centering
         \includegraphics[width=\textwidth]{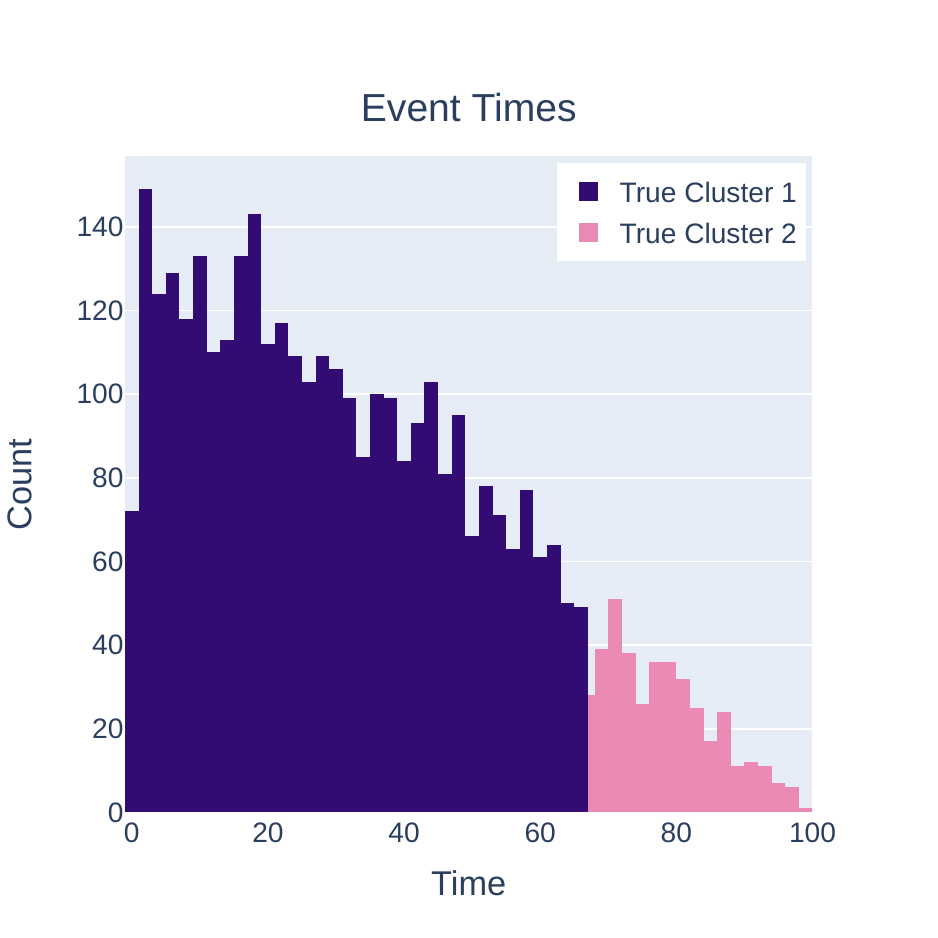}
         \caption{The event times of the training set colored by true cluster.}
         \label{fig: 1d survival clusters}
     \end{subfigure}
     \hfill
     \begin{subfigure}[b]{0.4\textwidth}
         \centering
         \includegraphics[width=\textwidth]{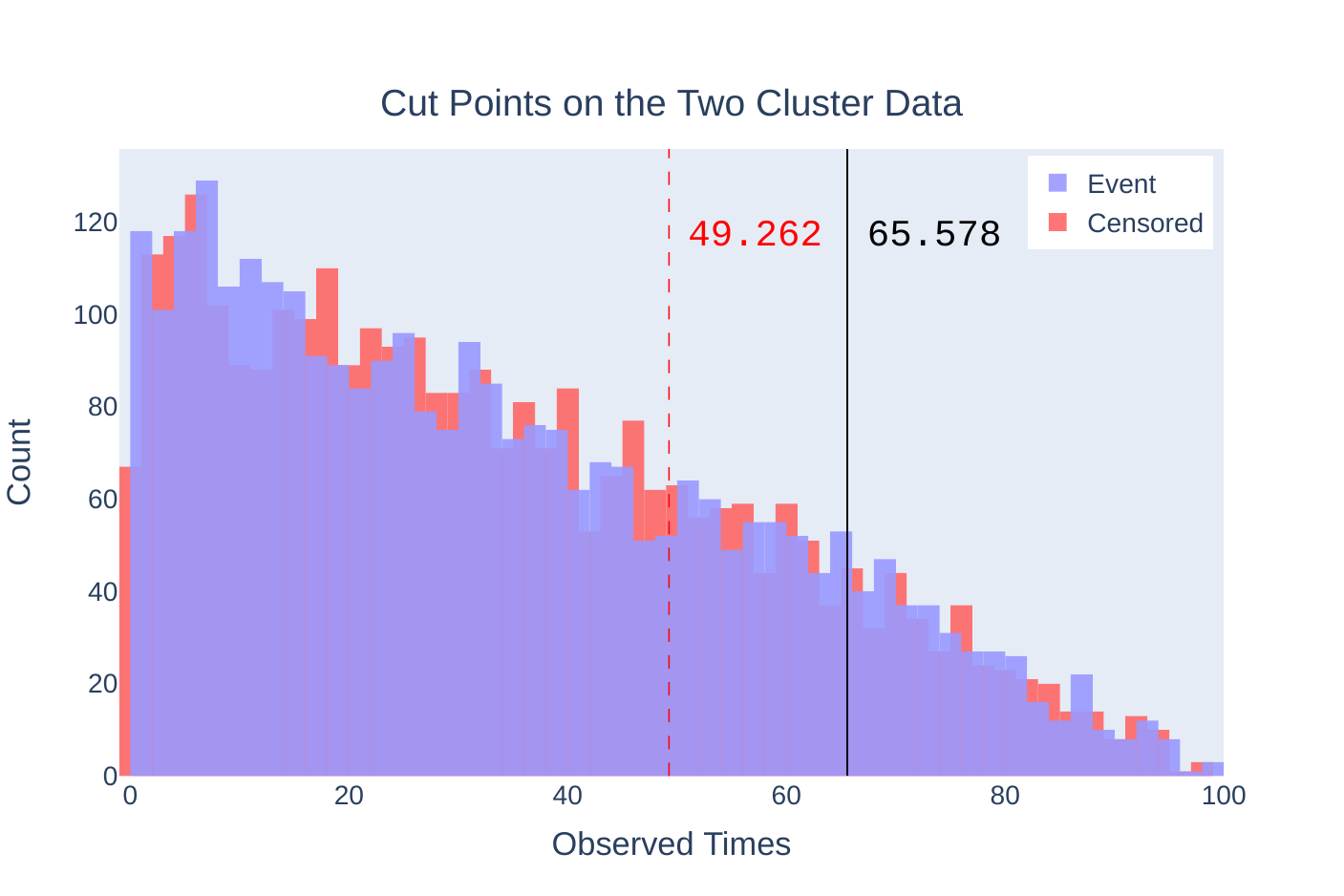}
         \caption{The DTNN Baseline (red, dashed) and learned (black, solid) cut point on the test set.}
         \label{fig: 1d learned}
     \end{subfigure}
        \caption{The event times and observed times of the two interval data. The true cut point is at time 67.}
        \label{fig: 1d}
\end{figure*}

We start with the simple case of data generated from two clusters with uniform censoring.
Cluster membership is generated using the \texttt{make\_moons} function \texttt{sklearn Python} package to get a noisy, nonlinear relationship between $p=2$ features \citep{scikit-learn}.
Figure \ref{fig: 1d feature clusters} shows the feature-cluster relationship; each cluster has $5,000$ observations for a total of $n=10,000$ observations.
These clusters are used to generate the event times.

Event times in Cluster 1 are generated uniformly on the interval $(0, 67]$ and event times in Cluster 2 are uniformly on the interval $(67, 100)$.
Censoring times are then generated uniformly throughout the entirety of $(0,100)$.
Note that while the censoring and event times are both uniformly distributed, the observed times are the minimum of the two and therefore not uniformly distributed.
These observed times are shown in Figure \ref{fig: 1d survival clusters}.
Because these intervals are determined by the relationship between the covariates, this set up simulates data generated with a true cut point at time 67.

Figure \ref{fig: 1d learned} shows out of sample test set along with the DTNN Baseline cut point in red at $t = 49.3$ and the learned cut point in black at $t=65.6$.
Knowing that the true cut point is at time 67 demonstrates the efficacy of our method.
Even with a starting point far from the true cut points, we are able to recover the true cut point.
Table \ref{table: syntehtic} reports the performance metrics, showing a large gain in CI.

In this simple example, many combinations of hyperparameters were able to recover the true cut point; reported are the results from using a small neural network with $h=32$ with Adam weight decay of 0 and a regularization strength of $\lambda = 1$.

\begin{table}[h]
    \centering
    \caption{Performance metrics for synthetic data. We report average metrics across 5-fold cross validation with standard errors in parentheses.}
    \begin{tabular}{r c c }
         &  Two Intervals & Four Intervals \\ \hline
         Learned CI &  {\bf 0.947 (0.001)} & {\bf 0.980 (0.012)} \\
         DTNN Baseline CI  & 0.797 (0.002) & 0.937 (0.007) \\
         % $\Delta\text{CI}$ & 0.15 & 0.043
         % Learned Entropy & 0.018 (0.0) & -2.95 (0.282)\\
         % Baseline Entropy & -0.204 (0.0) & -4.622 (0.289)\\
         % $\Delta\text{Entropy}$ & 0.222 & 1.672
    \end{tabular}
    \label{table: syntehtic}
\end{table}

\subsection{Learning Four Intervals}
\label{sec: 3d}
\begin{figure*}[h]
     \centering
     \begin{subfigure}[b]{0.27\textwidth}
         \centering
         \includegraphics[width=\textwidth]{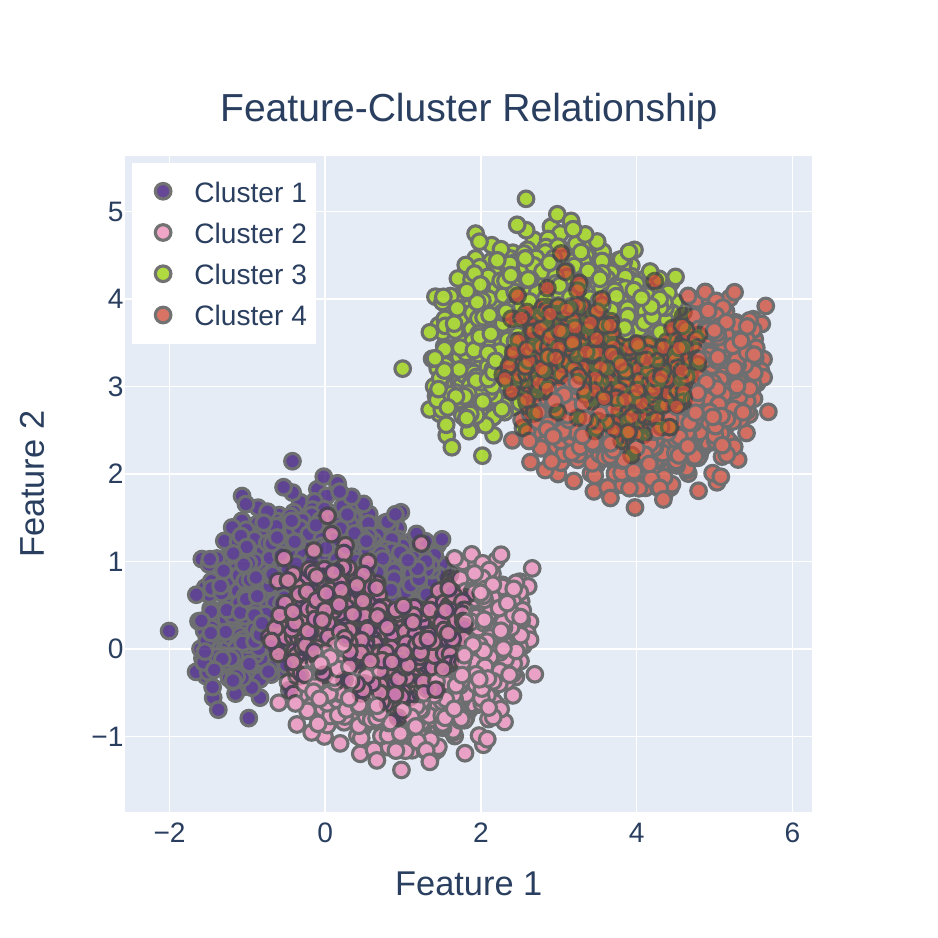}
         \caption{The noisy, nonlinear feature-cluster relationship.}
         \label{fig: 3d feature clusters}
     \end{subfigure}
     \hspace{2mm}
     % \\
     \begin{subfigure}[b]{0.27\textwidth}
         \centering
         \includegraphics[width=\textwidth]{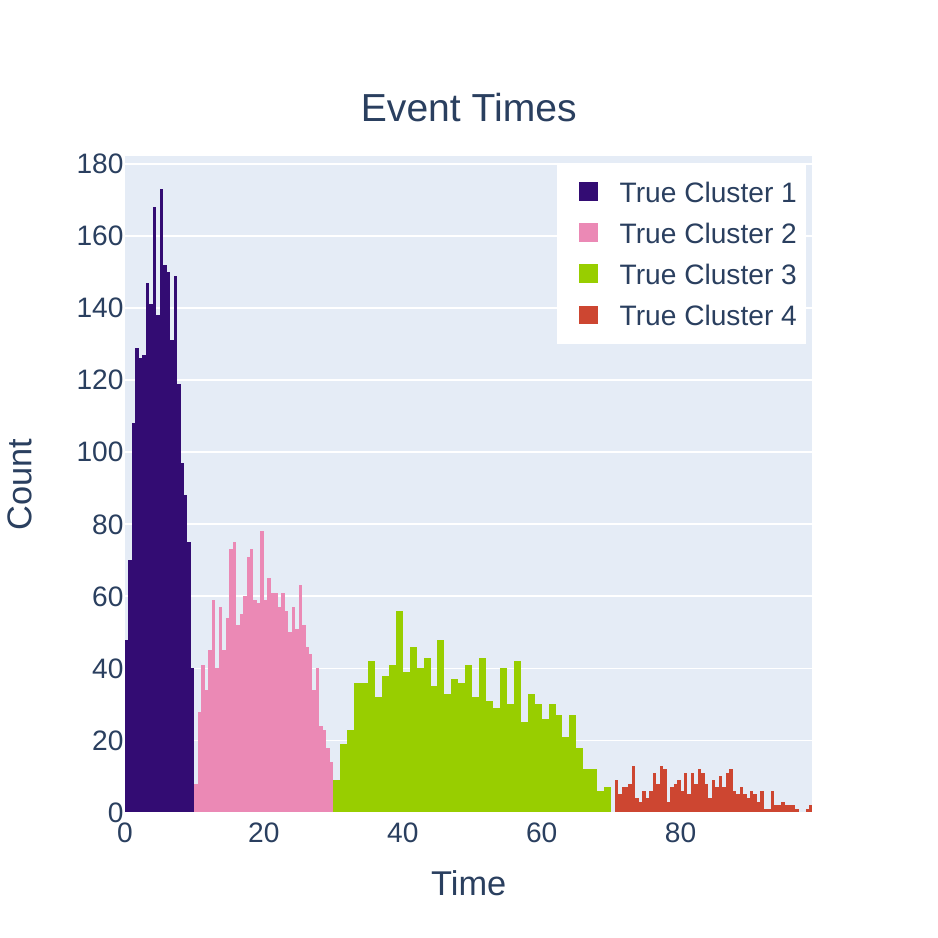}
         \caption{The event times of the training set colored by true cluster.}
         \label{fig: 3d clusters}
     \end{subfigure}
     \hspace{2mm}
     % \\
     \begin{subfigure}[b]{0.4\textwidth}
         \centering
         \includegraphics[width=\textwidth]{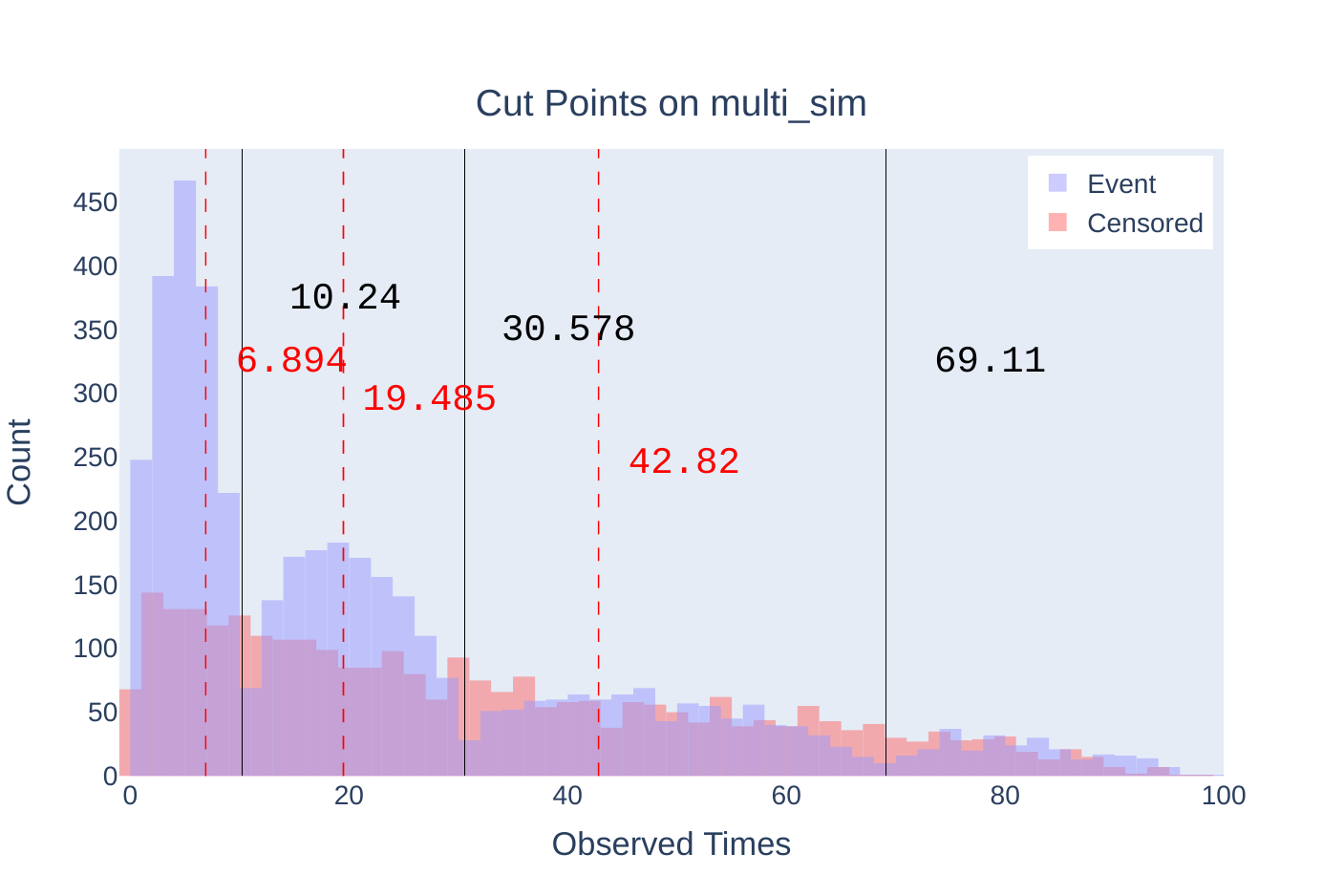}
         \caption{The DTNN Baseline (red, dashed) and learned (black, solid) cut point on the test set.}
         \label{fig: 3d learned}
     \end{subfigure}
        \caption{Event times and observed times of the four interval data. The true cut points are at 10, 30, and 70.}
        \label{fig: 3d}
\end{figure*}

With confidence in our ability to learn a single cut point when it is present in the data generation, we expand to learning three true cut points.
Again we use the \texttt{make\_moons} function to generate noisy, nonlinear relationships between $p=2$ features, however now for four separate clusters as shown in Figure \ref{fig: 3d feature clusters}; each cluster has $2,500$ observations for a total of $n=10,000$ observations.
Figure \ref{fig: 3d clusters} shows the how these clusters are used to generate event times.
Event times are generated using a $\text{Beta}(1.5, 1.5)$ distribution which are then scaled to be in the appropriate interval based on the observation's cluster.
The first cluster has observed times on the interval $(0,10]$, the second on the interval $(10, 30]$, the third on the interval $(30, 70]$, and the fourth on the interval $(70,100)$.
This corresponds to the true but points being at $t = 10, 30, 70$.
We again apply uniform censoring times to all observations.
Note that with uniform censoring, there are particularly few uncensored observations for events in the last interval.
This makes learning the final cut point more difficult.
% Here we try another naive baseline model by placing the baseline cut points evenly across the event time space at times $t = 25, 50, 75$.

Figure \ref{fig: 3d learned} shows that our method was able to successfully recover all three cut points despite the challenges due to censoring.
Table \ref{table: syntehtic} shows that the learned intervals provide an increase in CI over the DTNN Baseline.
Since we know the data generating mechanism, it is intuitive for this simulation example that including more than 3 cut points leads to worse performance as introducing more would overparametrize the model.
The results in the next section suggest that it is beneficial to consider models with fewer cut points even when the generating mechanism is unknown.

\section{Data Analysis}
\label{sec: data analysis}
\subsection{Real-World Data Sources}
\label{sec: Real-World Data Sources}
We apply our method to three real-world data sources of varying sizes. \\

\noindent
\textbf{German Breast Cancer Study Group (GBSG)} 
The GBSG data set is a publicly available data set introduced by \cite{gbsg}.
It is a multicenter clinical trial which includes $n=686$ patients with $p=8$ features.
The endpoint of recurrence free survival occurred for $299$ ($43.6$\%) patients.
\\ \\
\noindent
\textbf{Assay of Serum Free Light Chain (FL Chain)}
The FL Chain data set is a publicly available data set introduced by \cite{dispenzieri2012use} studies the relationship between nonclonal serum immunoglobulin free light chains and mortality.
We examine the data for the $n=6,524$ patients that had no missing data with $p=8$ features.
The endpoint of death occurred for $1,962$ ($30.1$\%) of these patients.
\\ \\
\noindent
\textbf{Pooled Stroke Risk Cohorts}
This is a combined dataset consisting of the Framingham Offspring Study \citep{feinleib1975framingham} ($n_{1} = 8,348$), The Atherosclerosis Risk in Communities Study \citep{aric1989atherosclerosis} ($n_{2} = 23,158$), and the Multi-Ethnic Study of Atherosclerosis ($n_{3} = 6,390$) \citep{bild2002multi}.
Data harmonization procedures and characteristics of the dataset have previously been described by \cite{hong2023predictive}.
We consider a total of $n=35,450$ data points of which $1,221$ ($3.44$\%) experience a stroke.
There are $p = 69$ features that include cardiovascular medical history, demographic indicators, and diet information.

\subsection{Results}
\begin{figure*}[hp]
     \centering
     \begin{subfigure}[b]{0.6\textwidth}
         \centering
         \includegraphics[width=\textwidth]{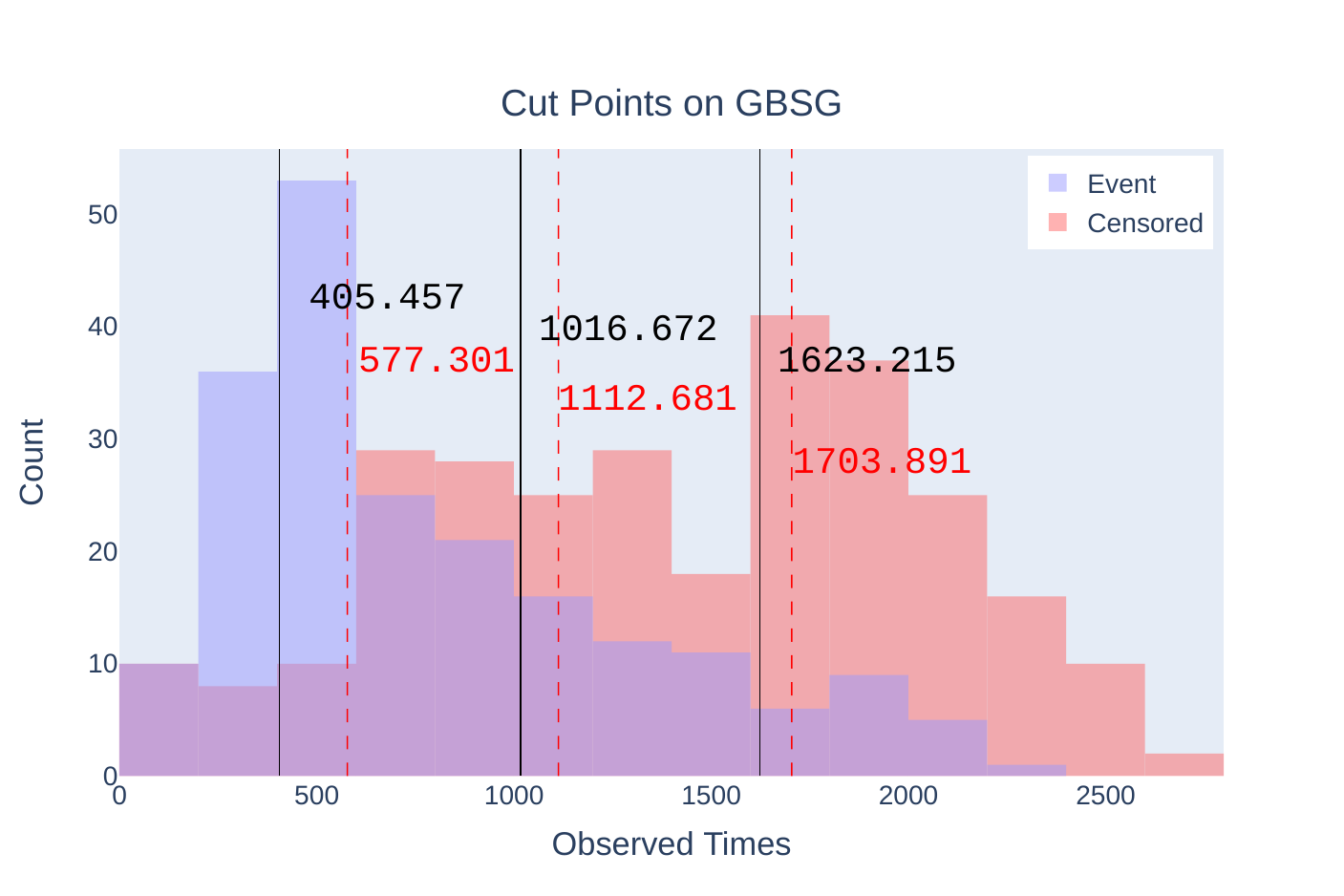}
         \caption{Cut points on GBSG data}
         \label{fig: gbsg}
     \end{subfigure}
     \\
     \begin{subfigure}[b]{0.6\textwidth}
         \centering
         \includegraphics[width=\textwidth]{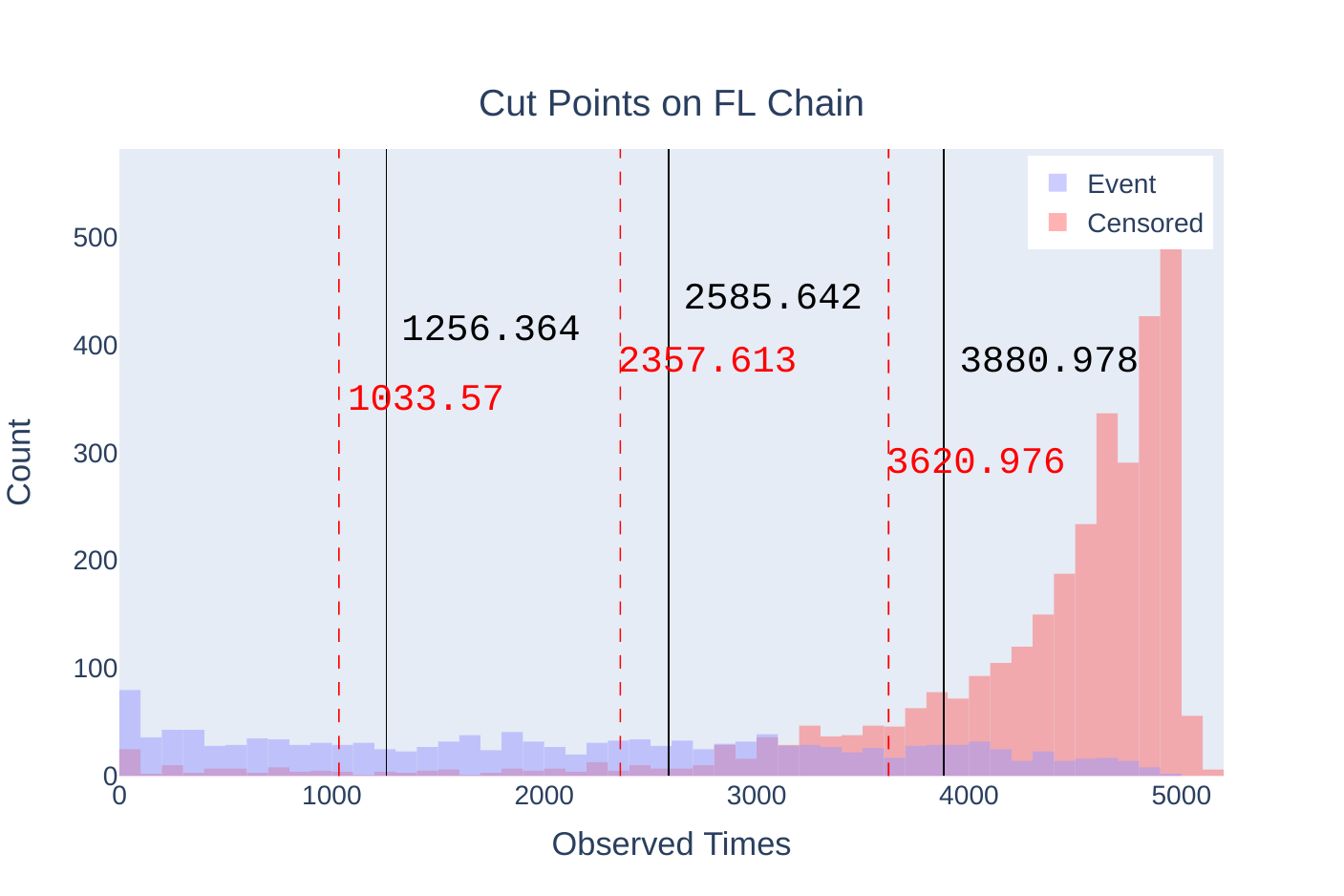}
         \caption{Cut points on FL Chain data.}
         \label{fig: fl chain}
     \end{subfigure}
     \\
     \begin{subfigure}[b]{0.6\textwidth}
         \centering
         \includegraphics[width=\textwidth]{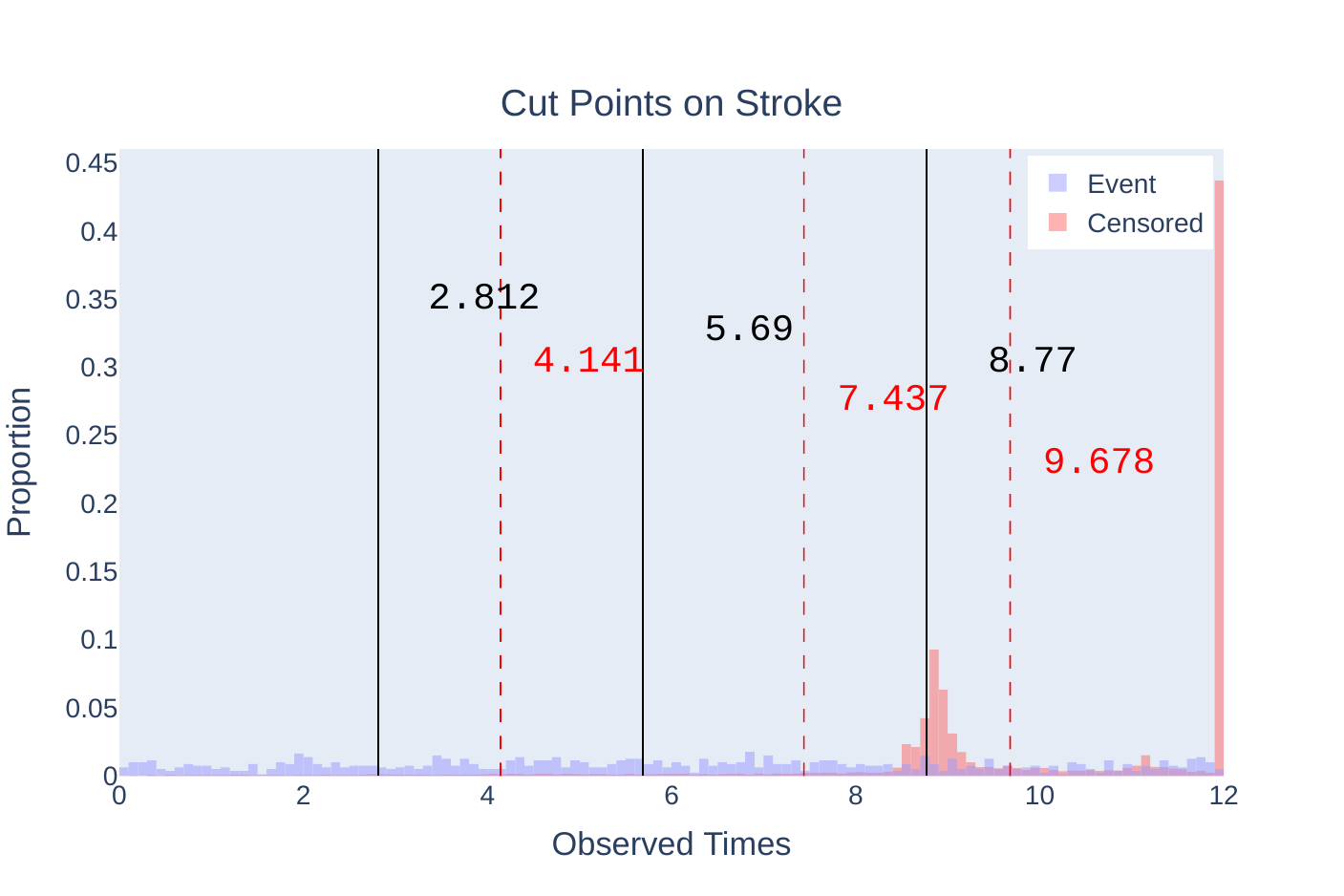}
         \caption{Cut points on Stroke data.}
         \label{fig: stroke}
     \end{subfigure}
        \caption{The DTNN Baseline (red, dashed) and learned (black, solid) cut points.}
        \label{fig: real data}
\end{figure*}

Figure \ref{fig: real data} shows the best learned cut points for each data set compared to the DTNN Baseline.
Note Figure \ref{fig: stroke} is a histogram of the proportion of observations rather than raw counts because of the high amount of censoring.
Table \ref{table: real data} shows the performance metrics for all real-world data sets.
The reported metrics are calculated on the held-out test sets not used for training or model validation.

Two interesting trends strongly support the benefits of our method. First, for all data sets, the CI was the highest for the models that used only 3 cut points and tended to decrease as more cut points were added.
Additionally, for all numbers of cut points, the predictive performance in both CI and AUC for the learned cut point model was better than the DTNN Baseline model.

Notice that the greatest improvement in CI was observed for the GBSG data set, which has the fewest observations among all data sets.
This underscores the importance of our proposed method.
For small data sets with a limited number of outcomes, it is necessary to limit the number of cut points, but performance can be improved by optimizing their locations. 
Notice that for the FL Chain data set, the DTNN Baseline model achieved its highest CI with 10 cut points, but this was still lower than the performance of the proposed approach using only 3 learned cut points.
Interestingly, the Stroke data set which had the most data points and the highest outcome imbalance also had a higher CI and AUC with 10 cut points than with 5.
Similar to the other data sets, it achieved its highest CI and AUC using 3 cut points.

Recall that a model that is well calibrated has a calibration slope near 1 and a calibration intercept near 0.
While the DTNN Baseline model had slightly better calibration slopes for the GBSG and FL Chain data sets for 3 cut points, the learned cut point model was better calibrated in nearly every setting with more cut points.
This demonstrates model robustness.

\begin{table*}[h!]
    \centering
    \caption{Test-set performance metrics for real-world data. Reported are average metrics across 5-fold cross validation and corresponding standard errors in parentheses.}
    \begin{tabular}{r l l l }
       &  3 Cut Points & 5 Cut Points & 10 Cut Points \\ \hline
       \textbf{GBSG} \\
          Learned CI & \textbf{0.744 (0.015)} &\textbf{0.68 (0.018)} & \textbf{0.671 (0.024)} \\
         DTNN Baseline CI  & 0.681 (0.027) & 0.651 (0.059) & 0.619 (0.065) \\
         % Learned Entropy & -0.305 (0.026) & -0.803 (0.059)  & -3.811 (4.307)  \\
         % Baseline Entropy & -0.361 (0.045)  & -0.656 (0.041) & -1.706 (0.084) \\ \hline
         % 
         Learned AUC & \textbf{0.804 (0.021)} &\textbf{ 0.801 (0.02)} & \textbf{0.822 (0.024)} \\
         DTNN Baseline AUC & 0.800 (0.03) & 0.750 (0.034) & 0.807 (0.016)\\
          Learned Calibration Slope & 0.813 (0.089) & 0.995 (0.112) & 0.793 (0.165) \\
         DTNN Baseline Calibration Slope & 1.00 (0.091) & 1.855 (0.154) & 1.397 (0.254)\\
         Learned Calibration Intercept & 0.178 (0.048) & 0.130 (0.042)  & 0.142 (0.06)\\
         DTNN Baseline Calibration Intercept & 0.129 (0.057) & -0.285 (0.084) & -0.215 (0.129) \\ 
 \hline
     \textbf{FL Chain} \\ 
        Learned CI & \textbf{0.798 (0.003)} &  \textbf{0.793 (0.003)} & \textbf{0.787 (0.004)}  \\
         DTNN Baseline CI  & 0.763 (0.007) & 0.772 (0.004) & 0.774 (0.012)\\
         % Learned Entropy & -0.068 (0.3) & -1.737 (0.071)   & -2.346 (0.624) \\
         % Baseline Entropy & -2.602 (0.074)  & -3.95 (0.065) & -8.905 (3.413) \\ \hline
         % 
         Learned AUC & \textbf{0.806 (0.004)} & \textbf{0.81 (0.004) }& \textbf{0.834 (0.002)}\\
         DTNN Baseline AUC & 0.788 (0.008) & 0.809 (0.004) & 0.828 (0.004)\\
         Learned Calibration Slope & 1.199 (0.103) & 1.182 (0.075) & 1.057 (0.098) \\
         DTNN Baseline Calibration Slope & 1.005 (0.04) & 0.999 (0.039) & 0.898 (0.076) \\
         Learned Calibration Intercept & 0.056 (0.007) & 0.050 (0.01)  & 0.085 (0.018) \\
         DTNN Baseline Calibration Intercept & 0.102 (0.008) & 0.100 (0.009) & 0.116 (0.026) \\ 
          \hline
      \textbf{Stroke} \\
        Learned CI & \textbf{0.789 (0.014)} & \textbf{0.747 (0.02)} & \textbf{0.765 (0.006)} \\
         Baseline CI  &  0.778 (0.01) & 0.739 (0.017)  & 0.758 (0.022) \\
         % Learned Entropy &  -0.751 (0.063) & -1.404 (0.028)  & -5.16 (0.743) \\
         % Baseline Entropy & -1.542 (0.029)  & -2.575 (0.075) &  -7.591 (0.344)
         Learned AUC &\textbf{ 0.766 (0.011)} & \textbf{0.701 (0.03)}& \textbf{0.723 (0.013) }\\
         DTNN Baseline AUC & 0.743 (0.01) & 0.681 (0.031)& 0.713 (0.021)\\
         Learned Calibration Slope & 0.783 (0.013) & 1.117 (0.195) &  1.270 (0.321) \\
         DTNN Baseline Calibration Slope &  1.098 (0.321) &  1.370 (0.248)& 1.295 (0.376) \\
         Learned Calibration Intercept & 0.019 (0.002) &  0.022 (0.003) & 0.025 (0.002) \\
         DTNN Baseline Calibration Intercept & 0.022 (0.003) & 0.019 (0.002) &  0.025 (0.002)
    \end{tabular}
    \label{table: real data}
\end{table*}

\newpage

\section{Discussion} 

Herein we develop a flexible method to learn an optimal partitioning of the event time space that does not place strong assumptions on the form of the event density. Our approach is designed for clinical applications in which it is advantageous to learn, from data, a time discretization that facilitates more accurate prediction.
The simulated examples demonstrated the ability of our method to recover cut points when they are truly present in the data generation mechanism.
Moreover, results on real data show that the approach improves prediction performance over otherwise equivalent, state of the art models that use a fixed discretization scheme.

Broadening this method could include considering higher dimensional output spaces and learning separating hyperplanes when multiple outcomes are of interest.
Another interesting extension motivated by the real-world data analysis would be to learn the number of cut points from the data instead of fixing it a priori.

\section*{Acknowledgements}
\label{sec:acknowledgements}
This study was supported by grant R61-NS120246 from the National Institute of Neurological Disorders and Diseases (NINDS). Jimmy Hickey's contribution to this work was funded by the T32 NIH grant number HL079896. Matthew Engelhard is supported by grant K01-MH127309 from the National Institute of Mental Health (NIMH).

The Framingham Heart Study is conducted and supported by the National Heart, Lung, and Blood Institute (NHLBI) in collaboration with Boston University (Contract No. N01-HC-25195 and HHSN268201500001I). This manuscript was not prepared in collaboration with investigators of the Framingham Heart Study and does not necessarily reflect the opinions or views of the Framingham Heart Study, Boston University, or NHLBI.

MESA and the MESA SHARe project are conducted and supported by the National Heart, Lung, and Blood Institute (NHLBI) in collaboration with MESA investigators. Support for MESA is provided by contracts N01-HC95159, N01-HC-95160, N01-HC-95161, N01-HC-95162, N01-HC-95163, N01-HC-95164, N01-HC-95165, N01-HC95166, N01-HC-95167, N01-HC-95168, N01-HC-95169 and CTSA UL1-RR-024156. This manuscript was not prepared in collaboration with MESA investigators and does not necessarily reflect the opinions or views of MESA, or the NHLBI.

The Atherosclerosis Risk in Communities study has been funded in whole or in part with Federal funds from the National Heart, Lung, and Blood Institute, National Institute of Health, Department of Health and Human Services, under contract numbers (HHSN268201700001I, HHSN268201700002I, HHSN268201700003I, HHSN268201700004I, and HHSN268201700005I). The authors thank the staff and participants of the ARIC study for their important contributions.

REGARDS is supported by cooperative agreement U01-NS041588 co-funded by the NINDS and the National Institute of Aging (NIA)

\bibliography{references}

\end{document}